\documentclass[12pt,letterpaper]{article}
\usepackage{authblk} 
\usepackage{amssymb}
\usepackage{amsmath}
\usepackage[colorlinks]{hyperref}
\usepackage{url}
\usepackage{array}
\usepackage{color,soul} 
\soulregister\ref{7} 
\soulregister\cite{7} 
\usepackage{float} 
\usepackage[colorinlistoftodos]{todonotes}
\usepackage{booktabs} 
\usepackage{rotating}
\usepackage{multirow}
\usepackage{transparent}
\usepackage{subcaption}
\usepackage{geometry}
\usepackage{layout}
\usepackage{bm} 
\def \Npatients {82}
\def \Ntrain {$\sim$60}
\def \Ntest {$\sim$20}
\newcommand\figscale{.99}

\renewcommand{\captionsize}{\footnotesize}
\begin{document} 
\clearpage
\newpage 
\title{Spatial Aggregation of Holistically-Nested Networks for Automated Pancreas Segmentation}
\author[1]{\small Holger R. Roth}
\author[1]{\small Le Lu}
\author[1]{\small Amal Farag}
\author[1]{\small Andrew Sohn}
\author[1]{\small Ronald M. Summers}
\affil[1]{\small Imaging Biomarkers and Computer-Aided Diagnosis Laboratory\\
Radiology and Imaging Sciences\\
Clinical Center\\
National Institutes of Health\\
Bethesda, MD 20892-1182, USA.}
\date{}
\maketitle
\begin{abstract}
Accurate automatic organ segmentation is an important yet challenging problem for medical image analysis. The pancreas is an abdominal organ with very high anatomical variability. This inhibits traditional segmentation methods from achieving high accuracies, especially compared to other organs such as the liver, heart or kidneys. In this paper, we present a holistic learning approach that integrates semantic mid-level cues of deeply-learned organ interior and boundary maps via robust spatial aggregation using random forest. Our method generates boundary preserving pixel-wise class labels for pancreas segmentation. Quantitative evaluation is performed on CT scans of \Npatients{} patients in 4-fold cross-validation. We achieve a (mean $\pm$ std. dev.) Dice Similarity Coefficient of 78.01\%$\pm$8.2\% in testing which significantly outperforms the previous state-of-the-art approach of 71.8\%$\pm$10.7\% under the same evaluation criterion.
\end{abstract}
\section{Introduction}
Pancreas segmentation in computed tomography (CT) is challenging for current computer-aided diagnosis (CAD) systems. While automatic segmentation of numerous other organs in CT scans such as liver, heart or kidneys achieves good performance with Dice Similarity Coefficients (DSC) of $>$90\% \cite{Wang2014Miccai,Chu2013Miccai,wolz2013automated}, the pancreas' variable shape, size and location in the abdomen limits segmentation accuracy to $<$73\% DSC being reported in the literature \cite{wolz2013automated,tong2015discriminative,okada2015abdominal,roth2015deeporgan}. Deep convolutional Neural Networks (CNNs) have successfully been applied to many high-level tasks in medical imaging, such as recognition and object detection \cite{yan2015bodypart}. The main advantage of CNNs comes from the fact that end-to-end learning of salient feature representations for the task at hand is more effective than hand-crafted features with heuristically tuned parameters \cite{zheng2015conditional}. Similarly, CNNs demonstrate promising performance for pixel-level labeling problems, e.g., semantic segmentation \cite{roth2015deeporgan,ronneberger2015unet,long2015fully,chen2014semantic}. Recent work in computer vision and biomedical imaging including fully convolutional neural networks (FCN) \cite{long2015fully}, DeepLab \cite{chen2014semantic} and U-Net \cite{ronneberger2015unet}, have gained significant improvements in performance over previous methods by applying state-of-the-art CNN based image classifiers and representation to the semantic segmentation problem in both domains. Semantic organ segmentation involves assigning a label to each pixel in the image. On one hand, features for classification of single pixels (or patches) play a major role, but on the other hand, factors such as edges (i.e., organ boundaries), appearance consistency and spatial consistency, could greatly impact the overall system performance \cite{zheng2015conditional}. Furthermore, there are indications of semantic vision tasks requiring hierarchical levels of visual perception and abstraction \cite{xie2015holistically}. As such, generating rich feature hierarchies for both the interior and the boundary of the organ could provide important ``mid-level visual cues'' for semantic segmentation. Subsequent spatial aggregation of these mid-level cues then has the prospect of improving semantic segmentation methods by enhancing the accuracy and consistency of pixel-level labeling.
\section{Methods}
In this work, we present a holistic semantic segmentation method for organ segmentation in CT which incorporates deeply learned organ interior and boundary mid-level cues with subsequent spatial aggregation. This approach to organ segmentation is completely based on machine-learning principles. No multi-atlas registration and label fusion methods are employed. Our methods are evaluated on CT scans of \Npatients{} patients in 4-fold cross-validation (instead of ``leave-one-patient-out'' evaluation often used in other work \cite{Wang2014Miccai,Chu2013Miccai,wolz2013automated}). 
\subsection{Candidate Region Generation}
\label{sec:region_candidates}
As a form of initialization, we employ a previously proposed method based on random forest (RF) classification \cite{roth2015deeporgan} to compute a candidate bounding box regions. We only operate the RF labeling at a low probability threshold of $>$0.5 which is sufficient to reject the vast amount of non-pancreas from the CT images. This initial candidate generation is sufficient to extract bounding box regions that completely surround the pancreases in all used cases with nearly 100\% recall. All candidate regions are computed during the testing phase of cross-validation (CV) as in \cite{roth2015deeporgan}. Note that candidate region proposal is not the focus of this work and assumed to be fixed for the rest of this study. This part could be replaced by other means of detecting an initial bounding box for pancreas detection, e.g., by RF regression \cite{criminisi2013regression} or sliding-window CNNs \cite{roth2015deeporgan}. 
\subsection{Semantic Mid-level Segmentation Cues}
We show that organ segmentation can benefit from multiple mid-level cues, like organ interior and boundary predictions. We investigate deep-learning based approaches to independently learn the pancreas' interior and boundary mid-level cues. Combining both cues via learned spatial aggregation can elevate the overall performance of this semantic segmentation system. Organ boundaries are a major mid-level cue for defining and delineating the anatomy of interest. It could prove to be essential for accurate semantic segmentation of an organ.
\subsubsection{Holistically-Nested Nets:} In this work, we explicitly learn the pancreas' interior and boundary image-labeling models via Holistically-Nested Networks (HNN). Note that this type of CNN architecture was first proposed by \cite{xie2015holistically} under the name ``holistically-nested edge detection'' as a deep learning based general image edge detection method. 
We however find that it can be a suitable method for segmenting the interior of organs as well (see Sec. \ref{sec:results}). HNN tries to address two important issues: (1) training and prediction on the whole image end-to-end (holistically) using a per-pixel labeling cost; and (2) incorporating multi-scale and multi-level learning of deep image features \cite{xie2015holistically} via auxiliary cost functions at each convolutional layer. HNN computes the image-to-image or pixel-to-pixel prediction maps (from any input raw image to its annotated labeling map), building on fully convolutional neural networks \cite{long2015fully} and deeply-supervised nets \cite{lee2014deeply}. The per-pixel labeling cost function \cite{long2015fully,xie2015holistically} offers the good feasibility that HNN/FCN can be effectively trained using only several hundred annotated image pairs. This enables the automatic learning of rich hierarchical feature representations (contexts) that are critical to resolve spatial ambiguity in the segmentation of organs. The network structure is initialized based on an ImageNet pre-trained VGGNet model \cite{simonyan2014very}. It has been shown that fine-tuning CNNs pre-trained on the general image classification task (ImageNet) is helpful to low-level tasks, e.g., edge detection \cite{xie2015holistically}.
\begin{figure}[htb]
\centering	\resizebox{\figscale\textwidth}{!}{\includegraphics{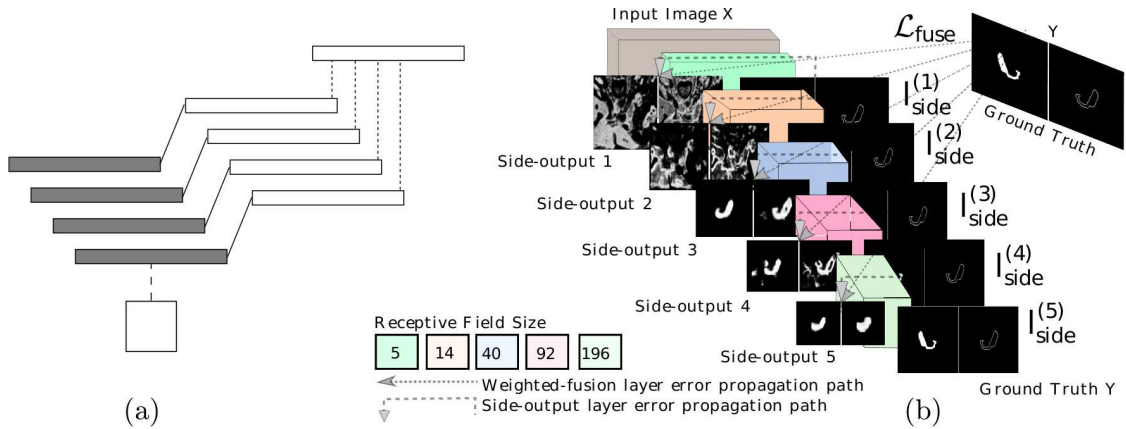}}  
	\caption{\captionsize Schematics of (a) the holistically-nested nets, in which multiple side outputs are added, and (b)  the \textbf{HNN-I/B} network architecture for both interior (left images) and boundary (right images) detection pathways. We highlight the error back-propagation paths to illustrate the deep supervision performed at each side-output layer after the corresponding convolutional layer. As the side-outputs become smaller, the receptive field sizes get larger. This allows HNN to combine multi-scale and multi-level outputs in a learned weighted fusion layer (Figures adapted from \cite{xie2015holistically} with permission).}
	\label{fig:hed}
\end{figure}
\subsubsection{Network formulation:} Our training data $S^{I/B} = \left\{(X_n , Y^{I/B}_n ), n = 1, \dots, N \right\}$ is composed of cropped axial CT images $X_n$ (rescaled to within $\left[0,\dots,255\right]$ with a soft-tissue window of $[-160, 240]$ HU); and $Y^I_n \in \left\{0,1\right\}$ and $Y^B_n \in \left\{0,1\right\}$ denote the (binary) ground truths of the interior and boundary map of the pancreas, respectively, for any corresponding $X_n$. Each image is considered holistically and independently as in \cite{xie2015holistically}. The network is able to learn features from these images alone from which interior (\textbf{HNN-I}) boundary (\textbf{HNN-B}) predication maps can be produced.

HNN can efficiently generate multi-level image features due to its deep architecture. Furthermore, multiple stages with different convolutional strides can capture the inherent scales of (organ edge/interior) labeling maps. However, due to the difficulty of learning such deep neural networks with multiple stages from scratch, we use the pre-trained network provided by \cite{xie2015holistically} and fine-tuned to our specific training data sets $S^{I/B}$ with a relatively smaller learning rate of $10^{-6}$. We use the HNN network architecture with 5 stages, including strides of 1, 2, 4, 8 and 16, respectively, and with different receptive field sizes as suggested by the authors\footnote{\scriptsize\url{https://github.com/s9xie/hed}.}.

In addition to standard CNN layers, a HNN network has $M$ side-output layers as shown in Fig. \ref{fig:hed}. These side-output layers are also realized as classifiers in which the corresponding weights are $\bm{\mathrm{w}} = (\bm{\mathrm{w}}^{(1)},\dots ,\bm{\mathrm{w}}^{(M )})$. For simplicity, all standard network layer parameters are denoted as $\bm{\mathrm{W}}$. Hence, the following objective function can be defined\footnote{\scriptsize We follow the notation of \cite{xie2015holistically}.}:
\begin{equation}\small
	\mathcal{L}_\mathrm{side}(\bm{\mathrm{W}},\bm{\mathrm{w}}) = \sum^{M}_{m=1}{\alpha_m}{l_\mathrm{side}^{(m)}}(\bm{\mathrm{W}},\bm{\mathrm{w}}^{m}).
	\label{equ:hed_loss}
\end{equation}
Here, $l_\mathrm{side}$ denotes an image-level loss function for side-outputs, computed over all pixels in a training image pair $X$ and $Y$. Because of the heavy bias towards non-labeled pixels in the ground truth data, \cite{xie2015holistically} introduces a strategy to automatically balance the loss between positive and negative classes via a per-pixel class-balancing weight $\beta$. This allows to offset the imbalances between edge/interior ($y=1$) and non-edge/exterior ($y=0$) samples. Specifically, a class-balanced cross-entropy loss function can be used in Equation \ref{equ:hed_loss} with $j$ iterating over the spatial dimensions of the image:
\begin{multline}\small
	l^{(m)}_\mathrm{side}(\bm{\mathrm{W}}, \bm{\mathrm{w}}^{(m)}) = - \beta \sum_{j\in Y_+}\log Pr\left(y_j=1|X;\bm{\mathrm{W}},\bm{\mathrm{w}}^{(m)}\right) - \\
	(1-\beta)\sum_{j\in Y_-}\log Pr\left(y_j=0|X;\bm{\mathrm{W}},\bm{\mathrm{w}}^{(m)}\right).
	\label{equ:hed_loss_balanced}
\end{multline}
Here, $\beta$ is simply $|Y_-|/|Y|$ and $1-\beta = |Y_+|/|Y|$, where $|Y_-|$ and $|Y_+|$ denote the ground truth set of \textit{negatives} and \textit{positives}, respectively. The class probability $Pr(y_j=1|X;W,w^{(m)}) = \sigma(a_j^{(m)}) \in [0,1]$ is computed on the activation value at each pixel $j$ using the sigmoid function $\sigma(.)$. Now, organ edge/interior map predictions $\hat{Y}ˆ{(m)}_\mathrm{side} = \sigma(\hat{A}ˆ{(m)}_\mathrm{side})$ can be obtained at each side-output layer, where $\hat{A}ˆ{(m)}_\mathrm{side} \equiv\{a_j^{(m)}, j = 1,\dots,|Y|\}$ are activations of the side-output of layer $m$. Finally, a ``weighted-fusion'' layer is added to the network that can be simultaneously learned during training. The loss function at the fusion layer $L_\mathrm{fuse}$ is defined as
\begin{equation}\small
	\mathcal{L}_\mathrm{fuse}(\bm{\mathrm{W}}, \bm{\mathrm{w}}, \bm{\mathrm{h}}) = Dist\left(Y, \hat{Y}_\mathrm{fuse}\right), 
\end{equation}
where $\hat{Y}_\mathrm{fuse}\equiv \sigma \left(\sum^M_{m=1} h_m^{\hat{A}_\mathrm{side}}   \right)$  with $h = \left(h_1,\dots,h_M\right)$ being the fusion weight. $Dist(.,.)$ is a distance measure between the fused predictions and the ground truth label map. We use cross-entropy loss for this purpose. Hence, the following objective function can be minimized via standard stochastic gradient descent and back propagation:
\begin{equation}\small
	(\bm{\mathrm{W}}, \bm{\mathrm{w}}, \bm{\mathrm{h}})^\star = \mathrm{argmin}\left(\mathcal{L}_\mathrm{side}(\bm{\mathrm{W}}, \bm{\mathrm{w}}) + \mathcal{L}_\mathrm{fuse}(\bm{\mathrm{W}}, \bm{\mathrm{w}}, \bm{\mathrm{h}})\right)
\end{equation}
\paragraph{Testing phase:} Given image $X$, we obtain both interior (HNN-I) and boundary (HNN-B) predictions from the models' side output layers and the weighted-fusion layer as in \cite{xie2015holistically}:
\begin{equation}\small
	\left(\hat{Y}^{I/B}_\mathrm{fuse}, \hat{Y}_\mathrm{side}^{I_1/B_1)}, \dots, \hat{Y}_\mathrm{side}^{I_M/B_M}\right) =  \bm{\mathrm{HNN-I/B}}\left(X, (\bm{\mathrm{W}}, \bm{\mathrm{w}}, \bm{\mathrm{h}}) \right).
\end{equation}
\subsection{Learning Organ-specific Segmentation Object Proposals}
\begin{figure}[htb]
\centering	\resizebox{\figscale\textwidth}{!}{\includegraphics{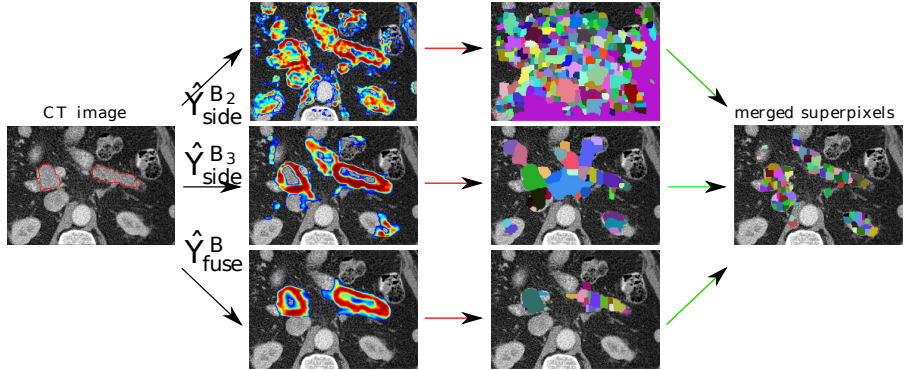}}
	\caption{\captionsize ``Multiscale Combinatorial Grouping'' (MCG) \cite{pont-tuset2015mcg} on three different scales of learned boundary predication maps from \textbf{HNN-B}: $\hat{Y}^{B_2}_{\mathrm{side}}$, $\hat{Y}^{B_3}_{\mathrm{side}}$, and $\hat{Y}^{B}_{\mathrm{fuse}}$ using the original CT image as input (shown with ground truth delineation of pancreas). MCG computes superpixels at each scale and produces a set of merged superpixel-based object proposals. We only visualize the boundary probabilities $p>10\%$.}
	\label{fig:mcg_superpixels}
\end{figure}

``Multiscale Combinatorial Grouping'' (MCG\footnote{\scriptsize\url{https://github.com/jponttuset/mcg}.}) \cite{pont-tuset2015mcg} is one of the state-of-the-art methods for generating segmentation object proposals in computer vision. We utilize this approach to generate organ-specific superpixels based on the learned boundary predication maps \textbf{HNN-B}. Superpixels are extracted via continuous oriented watershed transform at three different scales $(\hat{Y}_\mathrm{side}^{B_2}, \hat{Y}_\mathrm{side}^{B_3} , \hat{Y}_\mathrm{fuse}^{B})$ supervisedly learned by \textbf{HNN-B}. This allows the computation of a hierarchy of superpixel partitions at each scale, and merges superpixels across scales thereby, efficiently exploring their combinatorial space \cite{pont-tuset2015mcg}. This, then, allows MCG to group the merged superpixels toward object proposals. We find that the first two levels of object MCG proposals are sufficient to achieve $\sim88\%$ DSC (see Table \ref{tab:results} and Fig. \ref{fig:mcg_superpixels}), with the optimally computed superpixel labels using their spatial overlapping ratios against the segmentation ground truth map. All \textit{merged} superpixels $\mathcal{S}$ from the first two levels are used for the subsequently proposed spatial aggregation of \textbf{HNN-I} and \textbf{HNN-B}.
\subsection{Spatial Aggregation with Random Forest}
We use the superpixel set $\mathcal{S}$ generated previously to extract features for spatial aggregation via random forest classification\footnote{\scriptsize Using MATLAB's TreeBagger() class.}. Within any superpixel $s \in \mathcal{S}$ we compute simple statistics including the 1st-4th order moments and 8 percentiles $[20\%, 30\%, \dots, 90\%]$ on CT, \textbf{HNN-I}, and \textbf{HNN-B}. Additionally, we compute the mean $x$, $y$, and $z$ coordinates normalized by the range of the 3D candidate region (Sec. \ref{sec:region_candidates}). This results in 39 features describing each superpixel and are used to train a random forest classifier on the training \textit{positive} or \textit{negative} superpixels at each round of 4-fold CV. 
Empirically, we find 50 trees to be sufficient to model our feature set. A final 3D pancreas segmentation is simply obtained by stacking each slice prediction back into the space of the original CT images. No further post-processing is employed and spatial aggregation of \textbf{HNN-I} and \textbf{HNN-B} maps for superpixel classification is already of high quality. This complete pancreas segmentation model is denoted as \textbf{HNN-I/B-RF} or \textbf{HNN-RF}. 
\section{Results \& Discussion}
\subsubsection{Data:}
Manual tracings of the pancreas for \Npatients{} contrast-enhanced abdominal CT volumes were provided by a publicly available dataset\footnote{\scriptsize\url{http://dx.doi.org/10.7937/K9/TCIA.2016.tNB1kqBU}.} \cite{roth2015deeporgan}, for the ease of comparison. Our experiments are conducted on random splits of \Ntrain{} patients for training and \Ntest{} for unseen testing in 4-fold cross-validation. Most previous work \cite{Wang2014Miccai,Chu2013Miccai,wolz2013automated} use the leave-one-patient-out (LOO) protocol which is computationally expensive (e.g., $\sim15$ hours to process one case using a powerful workstation \cite{Wang2014Miccai}) and may not scale up efficiently towards larger patient populations.
\subsubsection{Evaluation:} 
\label{sec:results}
\begin{figure}[htb]
\centering	\resizebox{\figscale\textwidth}{!}{\includegraphics{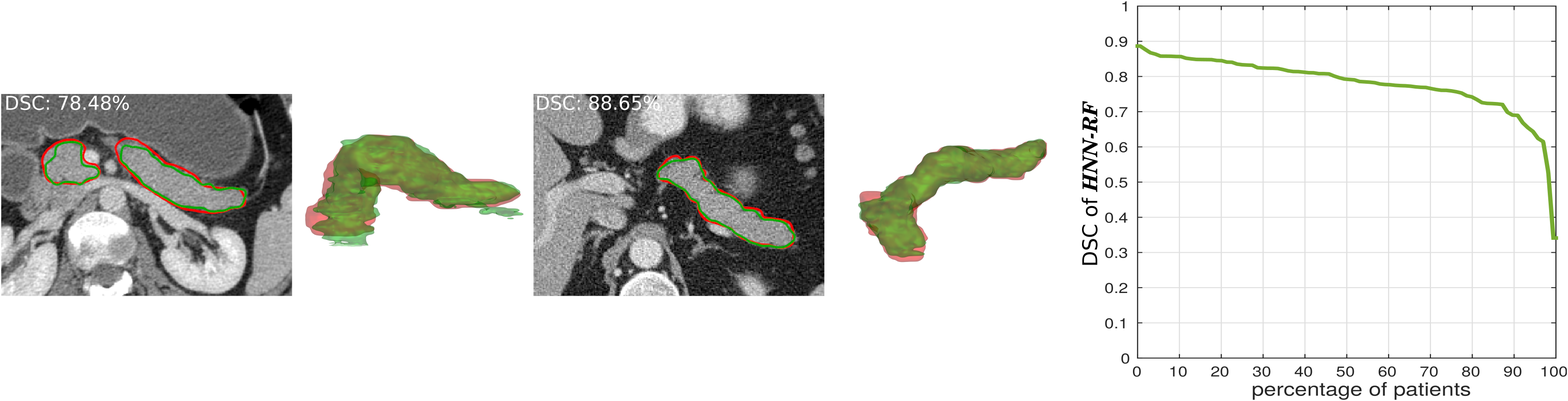}}
	\caption{\captionsize Examples of the \textbf{RF} pancreas segmentation (green) using the proposed approach in testing with the manual ground truth annotation (red). Case with DSC close to the data set mean and the maximum are shown. The percentange of total cases that lie  above a certain DSC with \textbf{RF} are shown on the right. 80\% of the cases achieve a minimum DSC of 74.13\%, and 90\% of the cases achieve a DSC of 69.0\% and higher.}
	\label{fig:axial_examples}
\end{figure}

Table \ref{tab:results} shows the improvement from \textbf{HNN-I} to using spatial aggregation via \textbf{HNN-RF} based on thresholded probability maps (calibrated based on the training data), using DSC and average minimum distance. The average DSC is increased from 76.99\% to 78.01\% statistically significantly (p$<$0.001, Wilcoxon signed-rank test). In contrast, using dense CRF (DCRF) optimization \cite{chen2014semantic} (with \textbf{HNN-I} as the unary term and the pairwise term depending on the CT values) as a means of introducing spatial consistency does not improve upon \textbf{HNN-I} noticeably (avg. DSC of 77.14\%, see Table \ref{tab:results}). To the best of our knowledge, our result comprises the highest reported average DSC (in testing folds) under the same 4-fold CV evaluation metric \cite{roth2015deeporgan}. Strict comparison to previous methods (except for \cite{roth2015deeporgan}) is not directly possible due to different datasets utilized. Our holistic segmentation approach with spatial aggregation advances the current state-of-the-art quantitative performance to an average DSC of 78.01\% in testing. To the best of our knowledge, this is the highest DSC ever reported in the literature. Previous state-of-the-art results range from $\sim$68\% to $\sim$73\% \cite{wolz2013automated,tong2015discriminative,okada2015abdominal}. In particular, DSC drops from 68\% (150 patients) to 58\% (50 patients) under the leave-one-out protocol as reported in \cite{wolz2013automated}. Our methods also perform with the better statistical stability (i.e., comparing 8.2\% versus 18.6\% \cite{Wang2014Miccai}, 15.3\% \cite{Chu2013Miccai} in the standard deviation of DSCs). The minimal DSC value is 34.11\% for \textbf{HNN-RF}, whereas \cite{Wang2014Miccai,Chu2013Miccai,wolz2013automated,roth2015deeporgan} all report patient cases with DSC $<$10\%. A typical patient result achieving a DSC close to the data set mean is shown in Fig. \ref{fig:axial_examples}.
Furthermore, we apply our trained \textbf{HNN-I} model on a different CT data set\footnote{\scriptsize 30 training data sets at \url{https://www.synapse.org/\#!Synapse:syn3193805/wiki/217789}.} with 30 patients, and achieve a mean DSC of 62.26\% without any re-training on the new data cases, but if we average the outputs of our 4 \textbf{HNN-I} models from cross-validation, we achieve 65.66\% DSC. This demonstrates that \textbf{HNN-I} may be highly generalizable in cross-dataset evaluation. Performance on that separated data will likely improve with further fine-tuning.
\begin{table}[htb]
\small
\centering
  \caption{\captionsize \textbf{4-fold cross-validation}: The DSC and average minimum distance (Dist) performance of our implementation of \cite{roth2015deeporgan}, optimally achievable superpixels, \textbf{HNN-I}, and \textbf{HNN-RF} spatial aggregation, and DCRF (best performance in bold).}
 \begin{subtable}[center]{.75\textwidth}
    \begin{tabular}{l|c|c|c|c|c}
    \toprule
		\toprule
    \textbf{DSC[\%]} & \cite{roth2015deeporgan}  & \textbf{Opt.} & \textbf{HNN-I} & \textbf{HNN-RF} & \textbf{DCRF}\\
		\midrule    
\textbf{Mean}	&71.42  &88.08  &	76.99  &	\textbf{78.01}  &	77.14  \\
\textbf{Std}      &10.11  &2.10    &	9.45    &	\textbf{8.20}    &	10.58  \\
\textbf{Min}	&23.99  &81.24  &	24.11  &	\textbf{34.11}  &	16.10  \\
\textbf{Max}	&86.29  &92.00  &	87.78  &	\textbf{88.65}  &	88.30  \\
    \bottomrule
		\bottomrule
    \end{tabular}%
 \end{subtable}%
\vskip 8pt    
 \begin{subtable}[center]{.75\textwidth}
 \raggedright
    \begin{tabular}{l|c|c|c|c|c}
    \toprule
		\toprule
    \textbf{Dist[mm]} & \cite{roth2015deeporgan}  & \textbf{Opt.} & \textbf{HNN-I} & \textbf{HNN-RF} & \textbf{DCRF}\\
		\midrule    
\textbf{Mean}	&1.53	&0.15	&0.70	&\textbf{0.60}	&0.69\\
\textbf{Std	}      &1.60	&0.08	&0.73	&\textbf{0.55}	&0.76\\
\textbf{Min}	&0.20	&0.08	&0.17	&\textbf{0.15}	&0.15\\
\textbf{Max}	&10.32	&0.81	&5.91	&\textbf{4.37}	&5.71\\
    \bottomrule
		\bottomrule
    \end{tabular}%
 \end{subtable}
  \label{tab:results}%
\end{table}%
\section{Conclusion}
In this paper, we present a holistic deep CNN approach for pancreas segmentation in abdominal CT scans, combining interior and boundary mid-level cues via spatial aggregation. Holistically-Nested Networks (\textbf{HNN-I}) alone already achieve good performance on the pixel-labeling task for segmentation. However, we show a significant improvement (p$<$0.001) by incorporating the organ boundary responses from the \textbf{HNN-B} model. \textbf{HNN-B} can improve supervised object proposals via superpixels and is beneficial to train \textbf{HNN-RF} that spatially aggregates information on organ interior, boundary and location. The highest reported DSCs of 78.01\%$\pm$8.2\% in testing is achieved, at the computational cost of 2$\sim$3 minutes, not hours as in \cite{Wang2014Miccai,Chu2013Miccai,wolz2013automated}. Our deep learning based organ segmentation approach could be generalizable to other segmentation problems with large variations and pathologies, e.g., tumors.
\paragraph{\footnotesize \textbf{Acknowledgments}} This work was supported by the Intramural Research Program of the NIH Clinical Center.
\footnotesize
\bibliographystyle{chicago} 
\bibliography{references_miccai2016}
\end{document}